\ifdefined\pdfoutput \pdfoutput=1 \fi  % arXiv: compile with pdflatex
\documentclass[11pt]{article}

\usepackage[margin=1in]{geometry}
\usepackage[utf8]{inputenc}
\usepackage[T1]{fontenc}
\usepackage{lmodern}
\usepackage{microtype}
\usepackage{booktabs}
\usepackage{longtable}
\usepackage{multirow}
\usepackage{graphicx}
\usepackage{xcolor}
\usepackage{amsmath}
\usepackage{url}
\usepackage[numbers,sort&compress]{natbib}
\usepackage[hidelinks]{hyperref}
\usepackage{iftex}
\ifPDFTeX
  \usepackage{CJKutf8}
  \newcommand{\zh}[1]{\begin{CJK}{UTF8}{bsmi}#1\end{CJK}}
\else
  \usepackage{xeCJK}
  \setCJKmonofont{Noto Sans Mono CJK TC}
  \newcommand{\zh}[1]{#1}
\fi

\newcommand{\cnum}[1]{\textcircled{\footnotesize #1}}

\usepackage{fancyhdr}
\fancypagestyle{preprint}{%
  \fancyhf{}%
  \fancyhead[L]{\footnotesize\textsc{Preprint}}%
  \fancyfoot[C]{\thepage}%
}

\newcommand{\benchname}{OpenTWBench}
\newcommand{\exambench}{\textsc{tw-exam-bench}}
\newcommand{\formosabench}{\textsc{tw-formosa-bench}}

\title{\benchname{} (Open Taiwan Benchmark): Taiwan-Domain Knowledge
Evaluation for Large Language Models, and How the Measurement Apparatus
Shapes the Score}

\author{%
  Liang-Hsun Huang$^{1,2,3}$ \quad Min-Yi Chen$^{1,2,3}$ \quad
  Chao-Chun Chuang$^{4,5}$\\[3pt]
  Yu-Wei Liu$^{6}$ \quad Che-Yu Wu$^{5}$ \quad Po-Feng Lee$^{7}$\\[5pt]
  $^1$Twinkle AI \qquad $^2$APMIC \qquad $^3$Open Taiwan Benchmark%
    \thanks{\emph{Open Taiwan Benchmark} is a non-profit organization that
    develops and maintains open, reproducible evaluation benchmarks for
    Taiwan-domain language models, including the \benchname{} suite presented
    in this paper.}\\[2pt]
  $^4$National Center for High-performance Computing (NCHC), Hsinchu, Taiwan\\[2pt]
  $^5$MonoLab \qquad $^6$Google Developer Expert, Cloud AI\\[2pt]
  $^7$Open Culture Foundation (OCF)
}

\date{}

\begin{document}
\maketitle
\thispagestyle{preprint}

\begin{abstract}
We present \benchname, an open evaluation suite for Taiwan-domain knowledge in
large language models, built around two complementary benchmarks.
\exambench{} contains 255{,}854 four-option questions across 148 academic
subjects, parsed from fifteen years of Taiwan's national examinations
published as open data by the Ministry of Examination; answer keys agree
across 31{,}312 duplicated question groups at 99.91\%, and every one of the 27
disagreements is accounted for. A reviewed second pass withdrew a further
2{,}180 items whose stems depend on a figure the text does not carry, so every
score we report is computed on the remaining 253{,}674. \formosabench{} contains 1{,}696
difficulty-curated questions about everyday and contemporary Taiwan --- the
Taiwan that licensing examinations never ask about --- generated from a
community-curated corpus under a construction discipline in which no model
ever writes a ground truth, and curated against a frozen seven-model pool so
that the pool scored at the passing margin (46.9--74.6\%, median
$\approx$54\%), with a leave-one-out experiment showing the selection
preserves model ranking at near-zero transfer bias. Within six weeks the
frontier had cleared that band: of the 108 models now on the public board, 22
score above its ceiling and the best scores 91.6\%.

Alongside the datasets, we document how much of a benchmark score is
contributed by the measurement apparatus rather than the model. On identical
questions and models, switching answer extraction from pattern matching to
explicit \verb|\boxed{}| moves accuracy from 20.8\% to 93.9\%; a single
placeholder character in a system prompt moved one model from last place
(58.2\%) to first (83.9\%) on the same benchmark; doubling the thinking
budget from 8k to 16k tokens bought 0.2\% while shuffling which
questions failed; and a three-judge validation panel proved to be one judge
in disguise, with two members approving 92--99\% of everything shown to them.
We release all datasets, per-item difficulty data, curation manifests, and a
public leaderboard at \url{https://opentwbench.ai}.
\end{abstract}

\section{Introduction}
\label{sec:intro}

Traditional Chinese as written in Taiwan is not a font variant of Chinese; it
is a linguistic and institutional context of its own, with its own calendar
(the Minguo era), legal system, medical licensing regime, place names,
vocabulary, and everyday culture. Language models are routinely deployed for
Taiwanese users, yet the evaluations behind those deployments are mostly
translations of English benchmarks or Simplified-Chinese suites, and the few
Taiwan-native benchmarks that exist are small, static, and increasingly
saturated. This paper describes \benchname, an open evaluation suite for
Taiwan-domain knowledge, and --- with equal emphasis --- what we learned about
how much of a reported benchmark number is produced by the measurement
apparatus rather than by the model being measured.

\paragraph{Two complementary benchmarks.} \exambench{} asks what national
examinations ask: 255{,}854 four-option questions across 148 academic
subjects, parsed from fifteen years of open-data papers published by Taiwan's
Ministry of Examination, with official answer keys whose internal consistency
we can measure (99.91\% agreement across 31{,}312 duplicated question groups,
every one of the 27 disagreements manually attributed;
Section~\ref{sec:keys}). \formosabench{} asks what examinations never ask:
1{,}696 closed-book questions about everyday and contemporary Taiwan ---
food, transit rules, public figures, local institutions --- generated from a
community-curated corpus under a construction discipline in which no model
ever writes a ground truth, then \emph{difficulty-curated} against a frozen
pool of seven contemporary open-weight models, so that the pool scored at the
passing margin (46.9--74.6\%, median $\approx$54\%). A leave-one-out
experiment shows the curation preserves model ranking with near-zero transfer
bias for mid-field models (Section~\ref{sec:curation}). The band held for
about six weeks; Section~\ref{sec:halflife} reports what the frontier did to
it, and what that says about how long any difficulty-targeted benchmark can
expect to hold.

\paragraph{The apparatus is part of the result.} Building the suite forced us
to measure things benchmark papers usually leave implicit, and the sizes are
uncomfortable. On identical questions and models, switching answer extraction
from pattern matching over free-form output to an explicit \verb|\boxed{}|
protocol moves accuracy from 20.8\% to 93.9\% --- a 73\% swing
contributed entirely by the harness. A single placeholder character in a
system prompt (\zh{甲} in the format example) moved one 675B model from last
place (58.2\%) to first place (83.9\%) on the same benchmark, because the
model copied the example's answer alphabet; the same prompt cost a different
model 17 questions out of 60{,}389. Doubling a reasoning model's thinking
budget from 8k to 16k tokens bought 0.21\% of accuracy at 1.72$\times$
the cost, while silently exchanging \emph{which} questions failed. And a
three-judge ``consensus'' validation panel turned out to be one judge in
disguise: two of its members approved 92--99\% of everything shown to them
(Section~\ref{sec:apparatus}). None of these effects are visible in a
leaderboard number; all of them are larger than typical claimed model
differences.

\paragraph{Contributions.}
\begin{enumerate}
\item \exambench: the largest Taiwan-native examination benchmark to date
  (255{,}854 items, 148 subjects, 15 years), with a measured key-consistency
  floor and a fully attributed conflict list.
\item \formosabench: a closed-book everyday-Taiwan benchmark with a
  zero-model-authored-ground-truth construction pipeline and a
  difficulty-curation method validated by leave-one-out transfer.
\item A quantified account of measurement-apparatus effects --- extraction
  protocol, prompt placeholders, thinking budget, judge-panel degeneracy ---
  each reproducible from released data.
\item A public leaderboard with per-item difficulty sidecars and frozen
  curation manifests, so every selection decision is auditable.
\end{enumerate}

All datasets, manifests, and evaluation code are released; the leaderboard
lives at \url{https://opentwbench.ai}.

\section{Related Work}
\label{sec:related}

\paragraph{Taiwan-native benchmarks.} TMMLU+ \citep{tam2024tmmluplus} is the
closest prior work: multiple-choice questions in Traditional Chinese drawn
largely from the same Ministry of Examination source we use, organized into
66 subjects. \benchname{} is best understood as a scaled-up successor on the
examination axis --- 255{,}854 items against TMMLU+'s $\sim$22k, 148 subjects
against 66, with fifteen years of coverage and a measured key-consistency
floor --- plus an everyday-knowledge axis (\formosabench) that examination
sources cannot provide at all.
\label{sec:tmmlu}
Because the two suites share a source, the same duplicate audit we run on
our corpus (Section~\ref{sec:keys}) applies to TMMLU+, and our build drops
2{,}948 items whose content fingerprints match TMMLU+ so that the suites
stay disjoint. Auditing TMMLU+ internally by order-insensitive content
matching finds 821 duplicate groups (1{,}720 of 23{,}072 rows); compared by
answer \emph{text} rather than letter --- option order differs across
appearances, and letter comparison manufactures 321 false conflicts ---
exactly one group truly disagrees, a very low label-error rate. More
consequentially, 431 of those groups span different splits, and 136 of the
20{,}160 test items appear verbatim in a non-test split (train, val, or the
few-shot dev pool) --- a contamination channel for anyone who trains or
few-shots on those splits and reports on test
(Appendix~\ref{app:tmmlu}). We release the audit script and matching lists.

TC-Eval \citep{hsu2023tceval} and the TAIDE bench cover Traditional Chinese
understanding at smaller scale; MT-Bench-TW and derivatives translate English
benchmarks, inheriting their topic distribution rather than Taiwan's.

\paragraph{Chinese-language suites.} C-Eval \citep{huang2023ceval} and CMMLU
\citep{li2023cmmlu} are Simplified-Chinese examination benchmarks for the PRC
context; their subject matter (PRC law, PRC accounting standards) and
orthography make them poor proxies for Taiwan deployment, and we treat
Simplified-character and mainland-vocabulary screening as a first-class filter
(Section~\ref{sec:formosa-construction}).

\paragraph{Examination-derived English benchmarks.} MMLU
\citep{hendrycks2021mmlu} established examination questions as an evaluation
substrate and also established their failure mode: label errors and
saturation \citep{gema2025mmluredux}. Our response to the error problem is to
measure it at the source --- duplicate answer keys across fifteen years give
an empirical noise floor ($\sim$$10^{-3}$) rather than an assumption --- and
our response to saturation is difficulty curation with transfer validation
rather than manual re-authoring.

\paragraph{Benchmark-measurement effects.} A growing literature documents
that scores depend on the harness: prompt-format sensitivity
\citep{sclar2024quantifying}, choice-ordering artifacts
\citep{zheng2024large}, and LLM-judge biases \citep{zheng2023judging}. Our
contribution here is not the phenomenon but its magnitude and specificity in
a CJK setting --- a single CJK placeholder character re-ranking a leaderboard,
extraction protocol dominating model choice by an order of magnitude --- with
paired same-model, same-item data released for each effect.

\section{\exambench: 255{,}854 Questions from Fifteen Years of National
Examinations}
\label{sec:exam}

\subsection{Source and extraction}

Taiwan's Ministry of Examination (\zh{考選部}) publishes every national
examination paper --- civil service, professional licensing, and technician
certification --- as open-data PDF files, together with the official answer
keys. We parse 15 years of these papers into a single benchmark following
the Twinkle Eval multiple-choice contract (\texttt{question},
\texttt{A}--\texttt{D}, \texttt{answer}), organized by \emph{academic
subject} rather than by profession: papers are grouped by what they examine
(pharmacology, constitutional law, railway engineering), not by which
professional category sat them. The suite covers 148 subjects across 19
fields, from medicine and nursing through maritime navigation to indigenous
affairs.

Extraction is deliberately conservative; a paper that cannot be parsed with
confidence is rejected rather than approximated. Three failure classes
consumed most of the engineering effort and are worth recording because each
would have silently corrupted the dataset:

\paragraph{Private-use glyphs.} The Ministry's PDFs encode circled option
markers as End-User Defined Characters in the Unicode Private Use Area
(U+E18C--U+E190); text extraction yields tofu unless these are mapped back to
A--E. Normalization must use NFC rather than NFKC, which would destroy other
distinctions the papers rely on.

\paragraph{Answer-grid alignment.} Official answer keys are laid out as
visual grids whose reading order does not match question order, and the first
cell of a row is frequently glued to its row label. Keys are therefore paired
to questions by page coordinates, with column tolerance derived from the
grid's own pitch. A one-cell shift here assigns every answer to its
neighbor --- the kind of error that is invisible in aggregate statistics and
fatal to a benchmark; it was caught by manually verifying individual
questions against domain knowledge.

\paragraph{Phantom questions.} Bare numbers inside question stems (statute
articles, dosages) can be mistaken for question numbers. Question numbering
is enforced to be monotonic, and any paper whose stated question count
(\zh{單選題數}, \zh{題數}) disagrees with the parsed count is rejected whole.

\subsection{Deduplication and key consistency}
\label{sec:keys}

Questions recur across years and examination categories. An
order-insensitive fingerprint (stem plus sorted option texts) identifies
31{,}312 duplicate groups. Crucially, duplicates are compared by \emph{answer
text}, not answer letter: option order varies across appearances, and
comparing letters manufactures false conflicts. (We initially made this
mistake, observing 322 apparent conflicts in an external dataset that reduced
to exactly one real conflict once compared by text; see
Section~\ref{sec:tmmlu}.)

Across all duplicate groups, official answer keys agree at
\textbf{99.9138\%}. All 27 flagged groups were manually attributed
(Appendix~\ref{app:conflicts}): 14 depend on figures absent from the
extracted text, 9 are detector artifacts (identical keys differing only in
whitespace or punctuation), one option set was legitimately rewritten
between years, one stem points at a case vignette outside the item, and
only \textbf{two} are true conflicts --- self-contained stems whose
official keys genuinely disagree. We quote the conservative detector-level
$99.9138\%$ as the noise floor a benchmark built from these papers
inherits; the attributed floor is an order of magnitude lower.

\subsection{Filtering}
\label{sec:filtering}

Four filters remove items that cannot be answered from text alone, each
calibrated against real false positives rather than assumed correct:
figure-dependent items (detected by deictic reference only --- a loose match
on \zh{圖} deletes library-science questions about \zh{圖書分類}); items
continuing a previous item (\zh{承上題}); English-language items, judged on
the stem only after two failed iterations that deleted 18{,}634 valid medical
questions containing English option terms; and items whose correct option is
pathologically longer than its distractors.

Table~\ref{tab:filters} gives the exact pipeline accounting, aggregated from
the per-subject build statistics carried on each dataset's card, so every row
can be re-derived from what is published.

\begin{table}[h]
\centering
\small
\begin{tabular}{lr}
\toprule
Pipeline stage & Count \\
\midrule
Exam papers seen / parsed & 13{,}812 / 11{,}621 \\
\quad rejected: incomplete parse / no questions / download failed & 2{,}002 / 188 / 1 \\
Raw items extracted & 629{,}343 \\
Folded as re-printings of an earlier year's item & 265{,}736 \\
\midrule
English-language items & 70{,}386 \\
Not four-choice & 14{,}779 \\
Figure-dependent & 6{,}113 \\
Officially voided (\zh{送分}) & 5{,}430 \\
Multi-select & 3{,}513 \\
Owned by an upstream benchmark (TMMLU+ overlap) & 2{,}948 \\
Continues a previous item (\zh{承上題}) & 2{,}876 \\
Bloated correct option & 1{,}198 \\
Duplicate options within an item & 326 \\
Stem too short & 132 \\
Cross-year answer-key conflict & 27 \\
No question text & 25 \\
\midrule
\textbf{Items published} & \textbf{255{,}854} \\
\midrule
Figure-dependent, second pass (Section~\ref{sec:secondpass}) & 2{,}180 \\
\midrule
\textbf{Items scored} & \textbf{253{,}674} \\
\bottomrule
\end{tabular}
\caption{\exambench{} build accounting across all 148 subjects. Rows below
the first rule are filtered items; folding of duplicate groups (31{,}312
groups across the 148 published subjects) is counted separately since those
items are re-printings, not defects. The published set is what the dataset ships; the scored set is what
every number in this paper is computed on, after the second-pass withdrawal
described in Section~\ref{sec:secondpass}.}
\label{tab:filters}
\end{table}

\subsection{A second pass the first one needed}
\label{sec:secondpass}

The deictic filter of Section~\ref{sec:filtering} removed 6{,}113
figure-dependent items, and we reported that as the figure problem solved. It
was not. Reviewing low-scoring subjects after the first release, we found
items whose stems depend on an absent figure without ever using a deictic
marker the filter could match --- \zh{圖中標示星號（*）之肌肉是}, where the
reference is a typographic symbol, or stems that describe a waveform or an
ultrasound image in prose and then ask what it shows. A reviewed pass over
all 148 subjects withdrew a further \textbf{2{,}180} items across 97
subjects: 85.8\% figure references, 7.0\% items continuing a question group,
and the remainder referencing an absent table, passage or attachment. The
withdrawals concentrate exactly where images carry the content --- dentistry
(277), electronics (166), diagnostic radiology (142) --- which is why
aggregate scores hid them.

Two things follow. Every score in this paper is computed on the remaining
253{,}674 items, and the withdrawal list ships with the dataset, so the
exclusion is auditable rather than a private correction.

The second is about how a filter gets validated.
Section~\ref{sec:filtering} measures what the figure rule wrongly deleted,
because a wrongly deleted item is visible: the library-science questions were
there to be counted. What the rule let through is not visible in the same
way --- nothing marks an item a filter failed to catch --- so the only way to
count those was to read all 148 subjects again. A filter checked on its
false-positive rate has been checked on the half of its errors that never
reach the published set; the second pass is the measurement the other half
requires. We would expect any corpus assembled this way to need one.
\section{\formosabench: Difficulty-Curated Everyday Taiwan}
\label{sec:formosa}

National examinations measure professional knowledge. They never ask who
designed the Weiwuying arts center, why a famous fish-head restaurant switched
to turkey during the foot-and-mouth outbreak, or what the MRT rules say about
drinking water. \formosabench{} covers this everyday, contemporary Taiwan:
closed-book, four-option questions derived from Taiwan.md, a community-curated
CC~BY-SA corpus of roughly 4.4M characters across 924 articles.

\subsection{Construction: no model writes a ground truth}
\label{sec:formosa-construction}

Items are drafted by a generator model (gpt-oss-120b), one per
paragraph-aligned window of the corpus (2{,}972 windows, median 1{,}119
characters, boundaries never mid-sentence). The prompt asks for facts
checkable outside the corpus and forbids document-local trivia. Quality then
comes from filters, not from trusting the generator:

\begin{itemize}
\item \textbf{Grounding.} Every keyed answer is verified against the window it
was written from by a judge that had previously caught 100\% of planted key
defects in a calibration study (Section~\ref{sec:judges}); 256 of 2{,}939
well-formed items failed and were dropped.
\item \textbf{Self-containment.} Stems that open on a pronoun
(\zh{她取得美國國籍的年份是？}) name nobody once the source is gone and are
rejected; pronouns \emph{inside} a stem resolve locally and are legal.
Stems referencing the passage (\zh{根據本文}) are stripped where the opening
is pure scaffolding and rejected where the reference is load-bearing.
\item \textbf{Anti-shortcut.} The generator placed the correct answer in
option A 84\% of the time; options are permuted by a hash of each item's
source window, after which ``always answer A'' scores 24.9\% and ``pick the
longest option'' 26.6\% (chance = 25\%).
\item \textbf{Orthography and vocabulary.} Zero Simplified-only characters and zero
mainland vocabulary written in Traditional characters (\zh{網絡},
\zh{軟件}), enforced by a lexicon whose strict terms fire zero times across
4.4M characters of Taiwanese prose and 255k examination questions ---
membership in the strict list requires exactly that.
\end{itemize}

An earlier design filtered instead for items \emph{unanswerable without the
source}. It kept registration fees and edition numbers --- corpus
memorization, not Taiwan knowledge --- and was discarded; the episode is a
useful caution that a filter optimizes exactly what it tests.

\subsection{Difficulty curation with transfer validation}
\label{sec:curation}

From 2{,}665 grounded items we publish the 1{,}696 that at least one model of
a frozen seven-model pool answers incorrectly (band $\text{pass} \le 6/7$;
pool, band and date frozen in a public manifest). On the pool this lands
scores at the passing margin: 46.9--74.6\%, median $\approx$54\%. That is a
statement about that pool on that date, not about the benchmark in
perpetuity; Section~\ref{sec:halflife} reports how long the band survived
contact with models released after it was frozen.

Pool-relative selection invites a fair objection: is the resulting set
adversarial to the pool, with scores that do not transfer to new models? We
answer it empirically with a leave-one-out experiment: select the band using
six models, then measure the held-out seventh on the selection.

\begin{table}[h]
\centering
\small
\begin{tabular}{lrrrrr}
\toprule
Held-out model & $n$ & Full set & Held-out & In-pool mean & Gap \\
\midrule
Mistral-Large-675B  & 1{,}679 & 83.9\% & 75.4\% & 54.7\% & $+20.7$ \\
gemma-4-31B         & 1{,}633 & 73.9\% & 61.3\% & 56.2\% & $+5.1$ \\
Gemma-3-27B         & 1{,}599 & 70.1\% & 56.3\% & 56.3\% & $+0.0$ \\
Nemotron-Super-120B & 1{,}643 & 70.9\% & 56.1\% & 57.2\% & $-1.2$ \\
gpt-oss-120b        & 1{,}622 & 69.3\% & 54.2\% & 57.1\% & $-2.9$ \\
Llama-3.3-70B       & 1{,}505 & 66.2\% & 52.9\% & 54.7\% & $-1.8$ \\
\bottomrule
\end{tabular}
\caption{Leave-one-out difficulty transfer. Each row selects the band
$\text{pass} \le 5/6$ using the other six models and measures the held-out
model on the resulting $n$ items. ``Full set'' is the held-out model's
accuracy on all 2{,}665 grounded items. Selection compresses scores by
8--16\%; mid-field models transfer with near-zero bias; stronger models
rise above the pool mean, as ability --- not selection artifact ---
predicts.}
\label{tab:loo}
\end{table}

Mid-field held-out models land within $\pm 3\%$ of the in-pool mean;
the two strongest held-out models exceed it by exactly the margin their
full-set superiority predicts, and the held-out ranking matches the full-set
ranking everywhere. As an additional validation with a model entirely outside
the pool and outside the generator's family, we evaluate Qwen3.8-27B, which
never participated in selection, under the identical protocol. It scores
72.0\% on the full 2{,}665 items --- mid-field, just under the pool's
full-set mean of 73.2\% --- and \textbf{59.0\%} [56.6--61.3] on the
published 1{,}696: a $13.0$-point compression squarely inside the
leave-one-out range, landing $+1.0$ points above the pool's curated-set
mean of 57.9\%, within the $\pm 3$-point transfer bias the leave-one-out
experiment predicts for mid-field models. Its nearest full-set neighbor
(gemma-4-31B, 73.9\% full) scores 59.0\% on the curated set --- a dead tie
with Qwen's 59.0\%, i.e.\ the adjacent ranking survives curation exactly.

\subsection{A note on fairness}
\label{sec:fairness}

The generator (gpt-oss-120b) also serves in the difficulty pool; its own
score on this benchmark carries a self-advantage and is annotated as such on
the leaderboard. The curation manifest and the per-item difficulty sidecar
are published so that any selection choice we made can be audited or redone.

\section{The Measurement Apparatus Shapes the Score}
\label{sec:apparatus}

Every result in this section holds the questions and the model constant and
varies only the apparatus. The effects are ordered by size.

\subsection{Extraction protocol: 20.8\% vs.\ 93.9\%}
\label{sec:extraction}

Twinkle Eval's \texttt{pattern} strategy sends the bare question with no
system prompt and pattern-matches an option letter out of free-form output;
its \texttt{box} strategy instructs the model to end with
\verb|\boxed{X}| and counts an unparseable response as wrong. On the same
civics subject and the same model, \texttt{pattern} reports 20.8\% and
\texttt{box} reports 93.9\%. The pattern strategy was not merely noisy: with
no stop discipline the model rambled past the answer, and the matcher once
extracted the letter ``W'' --- not an option --- from surrounding prose. The
73\% gap is larger than the gap between the best and worst model we
evaluate anywhere in this paper. Every score we publish uses \texttt{box},
and we publish parse rates next to accuracies so that ``the model got it
wrong'' and ``the harness could not read the answer'' remain separable.

\paragraph{What the leaderboard did instead, and why it was wrong.} Stating a
protocol is not the same as running one. Until 14 October 2026 the public
board scored each response with the \texttt{box} answer where the model
emitted one and a pattern match where it did not --- a fallback that reads a
bare leading letter, or failing that any option letter in the text. The
intent was charitable and the effect was regressive: the \emph{less} a model
complied with the answer format, the more of its score came from pattern
matching, so a model that never wrote \verb|\boxed{}| was scored entirely
leniently while one that always wrote it was scored entirely strictly. The
effect was not hypothetical: a model that wrote \verb|\boxed{}| on 6\% of
examination items sat two places above one that wrote it on 95\% and was
held to it, and a model that wrote it on 32\% sat six places above that same
model. 185 of the 215
published cells moved when we corrected this, 122 of them by more than a
point and 49 by more than ten; the median cell moved $-2.4$ points. The board
now publishes the \texttt{box} score, with the lenient figure beside it, so
that what a model knows and what format it writes it in are two columns
rather than one. We record the error rather than quietly repairing it: a
suite whose stated protocol and published numbers disagreed for six weeks is
evidence for this section's argument, not an exception to it.

\subsection{One placeholder character re-ranks the leaderboard:
58.2\% \texorpdfstring{$\to$}{->} 83.9\%}
\label{sec:placeholder}

Our first system prompt illustrated the format as \zh{例如} \texttt{\boxed{\text{\zh{甲}}}}
--- using \zh{甲}, the first of the traditional Chinese ordinals, as the
placeholder. Mistral-Large-675B copied the example's alphabet and answered
with \zh{甲乙丙丁} throughout: 827 of 828 ``unparsed'' responses on one
subject were in fact real answers in the wrong alphabet, and the model scored
58.2\% --- last place. Mapping the ordinals back to A--D predicted 83.4\%;
rewriting the prompt to name the alphabet (``one of A, B, C, D'') without an
example letter yielded a measured 83.9\% with a 100\% parse rate --- first
place among the pool at the time. The same defective prompt cost Qwen3.8-27B
only 17 responses out of 60{,}389.

The lesson is sharper than ``prompts matter'': the defect was
\emph{model-dependent}, so it did not shift the leaderboard --- it
\emph{re-ordered} it. A constant apparatus bias would cancel in comparisons;
a model-conditional one silently masquerades as capability difference. We
now treat any subject$\times$model cell with an outlier parse rate as an
apparatus alarm, not a model result.

\subsection{Thinking budget: doubling buys 0.21\%}
\label{sec:budget}

On accounting-audit-law, raising the generation cap from 8{,}192 to 16{,}384
tokens changed accuracy by $+0.21\%$ (CI $\pm 2.6\%$) at $1.72\times$ the
wall-clock cost, and unparsed responses fell only from 15.9\% to 14.0\%.
Item-level comparison shows why the aggregate is so stable: the larger budget
rescued 130 items and newly lost 103 --- the unparsed set \emph{churns}
rather than shrinks. Of responses that hit the cap, 97\% ran to the limit but
only 8\% were degenerate repetition loops; the dominant behavior is genuine
indecision, circling between statute readings without committing. Budget, in
other words, is not the binding constraint on hard legal items ---
decisiveness is.

\subsection{A judge panel that was one judge: 99\% / 92\% / 36\%}
\label{sec:judges}

An early Formosa validation pipeline used a three-judge panel with a
unanimity rule, which sounds rigorous. Measured per judge, two members
approved 99\% and 92\% of everything shown to them; the third approved 36\%.
``Three judges agree'' was operationally ``one judge decides.'' We replaced
the panel after a calibration study: ten candidate judges were shown items
with deliberately planted defects of three construction classes (wrong key,
ungrounded fact, malformed options). Every competent candidate caught 100\%
of planted defects; all disagreement between judges lay in unanchored
subjective dimensions (``is this interesting?''). The grounded-verification
judge used in Section~\ref{sec:formosa-construction} is the one that passed
this calibration --- and the general rule we draw is that a judge is only as
meaningful as the planted-defect study behind it.

\subsection{Answer-key consistency as a measured noise floor}

Section~\ref{sec:keys} treats the 99.9138\% cross-year key agreement not as a
quality boast but as an apparatus parameter: it bounds from below the error
any evaluation on this source inherits. Reporting model differences finer
than the source's own $\sim$$10^{-3}$ key noise, on subsets small enough for
that noise to bite, is measuring the apparatus again.

\section{Results}
\label{sec:results}

\subsection{\exambench}

\paragraph{Full-corpus anchor.} Qwen3.8-27B is our full-corpus anchor: it
answered all 148 subjects at a 16{,}384-token budget, scoring
\textbf{82.19\%} [82.04--82.34] over the 253{,}674-item cleaned canon --- the
255{,}854 published items less the 2{,}180 later withdrawn as unanswerable
without their missing figure --- at a 94.34\% parse rate.

\paragraph{The 54\% subject spread.} The aggregate hides the structure.
The model scores 97--98\% on clinical medicine subjects and 43--50\% on the
law cluster (accounting-audit law, tax law, civil procedure) --- a 54\%
spread within one model on one benchmark. Low parse rates (64--75\%)
co-occur with the law cluster: the same indecision documented in
Section~\ref{sec:budget} burns the budget without committing to an answer.
Subject-level scores are the unit at which this benchmark is informative;
we publish all 148.

\paragraph{Multi-model stratified sample.} Running 255{,}854 items per model
is impractical for a leaderboard, so the multi-model exam board uses a
deterministic stratified sample: per subject, $\min(150, n)$ items selected
by content hash (22{,}200 rows drawn, 22{,}021 after the withdrawals above),
with the full-corpus Qwen run as the anchor that calibrates
sample-to-population error.

Items are counted as rows, not as distinct question texts. Taiwan's
examinations reuse a stem verbatim across years with a fresh set of options
--- \zh{\texttt{下圖為何種通氣模式？}} is six different questions in
mechanical ventilation between 2014 and 2019, with three different answer
keys --- so 119 stems in the sample stand for 247 separate items. Keying
results on the stem would silently discard 128 of them; it is also why the
withdrawal list removes 179 rows where it names 174 stems.

The anchor calibrates that sample. Within a subject, items are selected by
content hash, which is independent of whether any model answers them
correctly; a sample drawn that way can differ from the population only
through composition. Weighting each subject's full-corpus accuracy by the
number of items the sample draws from it gives \textbf{80.1\%} against the
82.19\% population score --- a $-2.04$-point shift produced by the sample's
deliberate equal weighting of subjects, which moves weight away from the
large, high-scoring medical corpus (subjects with $\ge$2{,}000 items average
85.5\%; those under 500 average 77.4\%) toward the smaller legal and
administrative subjects where the model is weakest. Sample scores are
therefore comparable across models but sit systematically below what a
full-corpus run would report.

\begin{table}[h]
\centering
\small
\begin{tabular}{lrr}
\toprule
Model & Accuracy & \texttt{\textbackslash boxed\{\}} \\
\midrule
Mistral-Large-675B   & 83.4\% & 99.9\% \\
gemma-4-31B          & 82.1\% & 100.0\% \\
Qwen3.8-27B (anchor, composition-weighted) & 80.1\% & --- \\
gpt-oss-120b         & 77.1\% & 99.9\% \\
Nemotron-Super-120B  & 76.0\% & 89.2\% \\
Gemma-3-27B          & 66.7\% & 100.0\% \\
Llama-3.3-70B        & 66.7\% & 94.0\% \\
gemma-3-4B-T1-it     & 46.5\% & 95.4\% \\
Llama-3.2-3B-F1 (Instruct / Reasoning) & 45.4\% / 45.3\% & 98.5\% / 96.7\% \\
Llama-3.1-TAIDE-LX-8B & 16.5\% & 32.2\% \\
\bottomrule
\end{tabular}
\caption{\exambench{} stratified sample (22{,}021 rows), selected rows; the
full board of 107 models is published at \url{https://opentwbench.ai}. The
second column is the share of responses that contained a \texttt{\textbackslash boxed\{\}} at
all, and the accuracy is computed on those alone --- which is why
Llama-3.1-TAIDE-LX-8B, whose knowledge is comparable to the 3--4B models
above it, sits last: it wrote the requested format on under a third of
items. The anchor row places Qwen's full-corpus run on the sample's scale by
weighting its per-subject accuracies by the sample's per-subject counts; it
is an estimate on that scale, not a measurement taken on it (its full-corpus
figure is 82.2\%). Gemma-3-27B and Llama-3.3-70B differ by 0.06 points --- a
statistical tie. Scoring is row-wise over the 22{,}021 rows; 102 of the 106
sampled models answered all of them and four runs fell 9--36 rows short.}
\label{tab:examsample}
\end{table}

\subsection{\formosabench}

Scores of the frozen pool on the published (curated) benchmark:

\begin{table}[h]
\centering
\small
\begin{tabular}{lr}
\toprule
Model & Accuracy \\
\midrule
Mistral-Large-675B  & 74.6\% \\
gemma-4-31B         & 59.0\% \\
Nemotron-Super-120B & 54.3\% \\
Gemma-3-27B         & 53.1\% \\
gpt-oss-120b        & 51.8\%$^{\dagger}$ \\
Llama-3.3-70B       & 46.9\% \\
\bottomrule
\end{tabular}
\caption{Frozen-pool scores on \formosabench{} (1{,}696 items). Scores sit at
the passing margin by construction (Section~\ref{sec:curation}).
$^{\dagger}$gpt-oss-120b generated the draft items and its score carries a
self-advantage annotation.}
\label{tab:formosa-pool}
\end{table}

Qwen3.8-27B --- outside the pool, outside the generator family, never seen
during selection --- scores 59.0\% [56.6--61.3] on the published set
(72.0\% on the ungated 2{,}665), the fresh-model transfer validation
detailed in Section~\ref{sec:curation}. Across the frozen pool and this
held-out model --- the eight models the band was calibrated against ---
scores span 46.9--74.6\% with a median of 56.6\%: the passing margin the
curation targeted.

\begin{sloppypar}
Four Taiwan-tuned models land well above where their parameter count predicts
on the everyday-Taiwan axis (all served locally in bf16 --- there is no
quantization anywhere in the suite). Twinkle AI's gemma-3-4B-T1-it scores
\textbf{46.2\%} [43.9--48.6] on \formosabench{}, within a point of
Llama-3.3-70B's 46.9\% at $1/17$ the parameters; the Llama-3.2-3B-F1 pair
lands at 40.9\% (Instruct) and 39.0\% (Reasoning); TAIDE's
Llama-3.1-TAIDE-LX-8B-Chat scores \textbf{38.8\%} [36.5--41.1]. All four are
well clear of the 24.9\% shuffled-chance floor, and all four sit above where
3--8B parameters would put them.
\end{sloppypar}

The TAIDE figure needs its second column to be read at all. That model wrote
\verb|\boxed{}| on 67.0\% of \formosabench{} items and 32.2\% of examination
items, so both of its scores are bounded by its compliance rather than its
knowledge; held to the same pattern-matching extraction the rest of this
paragraph's models barely need, it scores 55.4\% and 47.9\%. The comparison
we can make without that caveat is among the compliant models, where
\verb|\boxed{}| rates run 83--99\%: a 4B Taiwan-tuned model is level with a
70B general model on everyday Taiwan while scoring 20 points below it on
professional examinations.

On \formosabench{} the reasoning-tuned F1 variant scores \emph{below} its
instruct sibling (39.0\% vs.\ 40.9\%) at an indistinguishable examination
score (45.3\% vs.\ 45.4\%) --- closed-book recall is not a reasoning problem,
and thinking longer cannot retrieve what the weights do not store (the same
mechanism as the budget result of Section~\ref{sec:budget}).

\subsection{Benchmark half-life: the band held six weeks}
\label{sec:halflife}

The difficulty band of Section~\ref{sec:curation} was calibrated against a
pool frozen on 24 August 2026. We report what happened next, because it
bounds how a difficulty-targeted benchmark should be read.

The public board carries 108 models on \formosabench{} as of 14 October
2026, including frontier systems released after the freeze. Their scores
span 0.0--91.6\% with a median of 45.4\%. Of those 108, only 29 fall inside
the 46.9--74.6\% band the curation targeted; 22 score above its ceiling, 8 of
them above 85\%, and the best scores 91.6\%. On \exambench{} the best model
scores 96.4\%. Seven weeks after freezing, in other words, the band no longer
describes the top of the field --- it describes the middle of it.

We do not read this as the curation having failed. The leave-one-out
experiment of Section~\ref{sec:curation} asked whether the selection
transfers to models \emph{outside the pool}, and it does: ranking is
preserved and mid-field transfer bias stays within $\pm 3$ points, which is
why the band still orders the 2026-vintage field correctly. What the band
cannot do is stay centred on a moving frontier, and no pool-relative
selection can. The honest form of the claim is therefore dated: these items
sat at the passing margin for the models available in August 2026, and the
difficulty sidecar is published per item so the band can be re-cut against a
newer pool without regenerating the benchmark.

The practical reading is that a difficulty-targeted benchmark has two
distinct lifetimes. Its \emph{discriminative} lifetime --- how long it
separates models --- is long: at the time of writing the published set still
spreads the field over 91 points, 66 of them between models that all write
the answer format and are separated by knowledge alone. Its
\emph{calibration} lifetime --- how
long a stated score band remains true --- was about six weeks. Papers that
report the second as though it were the first will be wrong within a
release cycle, and we would have been had we published this one in August
without dating the claim.

\subsection{What the two benchmarks disagree about}

Over the 107 models carrying both scores, the two axes agree far more than
they disagree: Spearman's $\rho = 0.97$. That figure is inflated by range ---
a board spanning 0\% to 96\% will correlate on almost any two measures of
capability --- but restricting it does not rescue the difference: $\rho$ is
0.89 over the top twenty models by examination score and 0.94 over the 27
scoring above 80\%. On the evidence, a second Taiwan benchmark buys very
little \emph{general} ranking information over the first.

What it buys is a specific kind of disagreement, and the size of it depends
on which extraction the question is asked under. Ranking each model on both
axes and taking the difference, the seven Taiwan-tuned models on the board
--- the TAIDE, Twinkle AI, Breeze and Llama-3-Taiwan families --- gain a mean
of \textbf{6.9} places on Formosa relative to their examination rank (median
$+3$), while the other 100 models move by a mean of $-0.5$ places (median
$0$). On the published \verb|\boxed{}| scores that is the whole effect, and
it is not a robust one: three of those seven models write the answer format
on under 7\% of items and sit near zero on both axes, carrying no signal in
either direction, and the mean is dominated by one model whose compliance
differs by a factor of two between the two benchmarks.

Holding extraction constant isolates the part of the effect that is about
knowledge. On the lenient figure published beside every score --- the same
responses, read by a pattern matcher instead of only inside
\verb|\boxed{}| --- the seven gain a mean of \textbf{15.6} places (median
$+17$) against $-1.2$ for the other 94, and the median model scores 7.1
points \emph{lower} on Formosa than on the examination while
Llama-3.1-TAIDE-LX-8B-Chat scores 7.5 points \emph{higher} and
gemma-3-4B-T1-it 3.7 points higher. We report both because they answer
different questions. What a leaderboard owes its readers is one stated
protocol applied to everyone, which is the \verb|\boxed{}| column. What a
claim about Taiwan-specific tuning needs is extraction held fixed, because
otherwise a model is being measured on a format convention that has nothing
to do with Taiwan.

Either way this is a sharper justification for two axes than ``the rankings
reorder,'' and a narrower one. The examination axis measures professional
knowledge and is well predicted by general capability; the Formosa axis
measures everyday Taiwan, where general capability predicts most of the
variance and Taiwan-specific tuning accounts for a systematic remainder ---
one that is clearly measurable with extraction fixed and partly obscured by
format compliance without it. A suite that carried only the examination would
rank the field almost identically, and would miss precisely the models built
for the thing it claims to measure.

\subsection{Taiwan-specific failure modes}
\label{sec:failures}

Reading failures, not just counting them, surfaces a taxonomy specific to
the Taiwan setting (drawn from structured-output evaluations during suite
development; the underlying benchmark is second-wave, but the failure classes
are observable in free-text answers as well):

\begin{itemize}
\item \textbf{Minguo-calendar arithmetic.} Given a bare Minguo year
(\zh{115年} without the \zh{民國} prefix), Llama-3.3-70B converts with a
fixed off-by-one (115 $\to$ 2027); Mistral-Large hallucinates an unrelated
date entirely, yet converts correctly whenever the \zh{民國} prefix is
present --- a one-token feature flips the behavior.
\item \textbf{CJK glyph corruption.} Llama-3.3-70B invents nonexistent
characters (\zh{左翐} for \zh{左營}) --- an orthography violation,
not a knowledge error, and one our filters treat as a specification
failure rather than an acceptable variant.
\item \textbf{Romanization drift.} TAIDE emits pinyin (\texttt{banqiao},
\texttt{zuoying}) where Chinese station names are required.
\item \textbf{Error detectability decides survival.} A corrupted station
name fails loudly and gets retried into correctness; a wrong-but-legal date
fails silently and survives to the final answer. Downstream validation is a
safety net only where errors are loud.
\end{itemize}

\section{Discussion}
\label{sec:discussion}

\paragraph{Quality cannot be requested, only filtered.} A recurring pattern
across the suite's construction: asking a generator model for a property
does not produce it. Explicitly prompting for corpus-independent questions
\emph{lowered} the corpus-independence rate (40\% $\to$ 32\% in our
generator trials); asking three judges for consensus produced one judge's
opinion. Every quality property we actually obtained came from a filter with
a measurable false-positive story --- grounding verification calibrated on
planted defects, shortcut baselines measured after permutation, screening
lexicons validated for zero false fires on 4.4M characters of native prose.
We suggest this as a design stance for benchmark construction generally:
generation proposes, filtering disposes.

\paragraph{Report the apparatus with the score.} The effects in
Section~\ref{sec:apparatus} range from 0.2\% to 73\%. Model cards and
leaderboards report the score; almost none report the extraction protocol,
the prompt's placeholder characters, the parse rate, or the judge approval
distribution --- parameters that our data show can dominate the ranking. Our
leaderboard publishes parse rate beside accuracy for every cell, and we
would argue any CJK benchmark should: the placeholder incident is invisible
without it.

\paragraph{Difficulty curation is a lens, not a distortion --- if validated.}
Selecting items a frozen pool finds hard invites the charge of building an
adversarial set. The leave-one-out experiment is the answer we owe:
ranking preserved, mid-field transfer bias within $\pm 3\%$, and
stronger models rising above the pool mean exactly as ability predicts. What
the experiment does not buy is permanence: a pool-relative band tracks the
pool, and Section~\ref{sec:halflife} measures how quickly a stated band stops
describing the top of the field. Curation choices should be published with
the date and pool attached, as ours are. The alternative interventions we
measured were far weaker: pruning all-correct
items moved top scores by only 0.6--3.0\% (the all-correct fraction is
small, and numerator and denominator shrink together). Scores are lowered by
writing harder items or selecting harder subsets --- not by deleting easy
ones.

\paragraph{Limitations.} (1) The difficulty pool is seven open-weight models
served through one provider, frozen in August 2026. The board has since grown
to 108 models and the frontier has cleared the band
(Section~\ref{sec:halflife}), which validates the selection's transfer but
also dates it: the published band reflects one pool at one moment, and
re-cutting it against a current pool is left to the per-item difficulty
sidecar rather than done here. (2) \formosabench's draft items come
from one generator; the zero-ground-truth discipline bounds the damage a
generator can do to \emph{correctness}, but topic distribution still
reflects its choices and the source corpus's coverage. (3) \exambench{}
inherits the Ministry's own $\sim$$10^{-3}$ key noise; we can measure it but
not remove it. (4) Both benchmarks are four-option multiple choice;
free-form and agentic Taiwan-domain evaluation (tool use with Minguo dates
and CJK arguments, long-context, instruction following) is built and
measured but deliberately held for a second release, where its own
apparatus story --- including a cross-leaderboard calibration showing our
first multi-turn design ran 2.4--7.6$\times$ easier than BFCL v3
\citep{patil2025bfcl} ---
deserves full treatment.

\paragraph{Knowledge first, then agency.} Most benchmark effort at the time
of writing goes to coding and agentic ability, and that is a reasonable place
for the field's attention --- but it is not a place a Taiwan-specific suite
can start. An agent booking a high-speed rail seat, reading a statute or
filling in a government form fails on a Minguo date or a corrupted station
name long before it fails on planning (Section~\ref{sec:failures}): those are
knowledge and orthography failures, and no coding benchmark will ever surface
them. \exambench{} and \formosabench{} are therefore the first two
instruments of an open Taiwan evaluation suite rather than the whole of it.
The agentic layer --- tool use with Minguo arguments and CJK parameters, long
context, instruction following in Traditional Chinese --- is built and
measured already, and follows in the second release. Open Taiwan Benchmark's
remit is that whole sequence, published on the same terms throughout: open
data, frozen manifests, dated pools, and the apparatus reported beside every
score.

\paragraph{Conclusion.} \benchname{} contributes the largest Taiwan-native
examination benchmark, a validated difficulty-curated everyday-knowledge
benchmark, and --- we would argue most durably --- a worked demonstration
that the measurement apparatus is a first-class experimental variable. The
datasets will age --- we measured how fast, and it was faster than we
expected --- but the discipline of publishing parse rates, curation
manifests, dated pools, withdrawal lists and planted-defect judge
calibrations alongside scores should not.

\section*{Acknowledgements}

Four kinds of help made this suite possible, and no two of them came from
the same place.

The \textbf{National Center for High-performance Computing} provided the B200
compute and the API access behind the leaderboard, the latter through
\textbf{Taiwan AI RAP} (\url{https://rap.genai.nchc.org.tw/}), which served
part of the board. Both halves mattered and
they are not interchangeable: the hosted endpoints reach the frontier models
that cannot be run locally at all, and the B200 capacity serves the large
open-weight models that no single workstation can hold. The full-corpus
anchor --- 253{,}674 items across 148 subjects --- and the 108-model board
are both downstream of that support.

The \textbf{Open Culture Foundation} sponsored the Apple Silicon Mac Studio
that serves every model our CUDA machine cannot. The limit there was not only
size: a 24\,GB consumer GPU capped at CUDA 12.5 cannot start a runtime new
enough for several recent architectures, whatever their parameter count, and
256\,GB of unified memory removes both constraints at once. The smaller
open-weight models on the board, down to the sub-billion ones, are evaluated
there under the identical protocol.

The \textbf{Google Developer Expert Program} provided the Gemini credits
behind the four Gemini models on the board. They are not a marginal
addition to it: they hold the first three places on both benchmarks, and the
91.6\% \formosabench{} ceiling and 96.4\% \exambench{} ceiling that
Section~\ref{sec:halflife} measures the difficulty band against are theirs.
A benchmark that cannot reach the frontier cannot report where the frontier
is, and the half-life result is the part of this paper that depended most
directly on being able to. The programme that provided them recognises
one of this paper's authors; we note that rather than leave it implicit, in
the same spirit as the Taiwan.md and generator annotations.

\formosabench{} exists because \textbf{Taiwan.md} exists: an openly licensed,
community-curated corpus of everyday and contemporary Taiwan, created and
maintained by Che-Yu Wu, broad enough that a benchmark about what the
examinations never ask could be built from it at all. We note the overlap
rather than leave it implicit --- the corpus one of this paper's authors
maintains is the corpus \formosabench{} is drawn from --- in the same spirit
as the generator annotation of Section~\ref{sec:fairness}. A benchmark can
only measure what somebody first took the trouble to write down, and that
corpus was years of work before there was an evaluation suite to use it.

\bibliographystyle{plainnat}
\bibliography{references}

\begin{thebibliography}{10}
\providecommand{\natexlab}[1]{#1}
\providecommand{\url}[1]{\texttt{#1}}
\expandafter\ifx\csname urlstyle\endcsname\relax
  \providecommand{\doi}[1]{doi: #1}\else
  \providecommand{\doi}{doi: \begingroup \urlstyle{rm}\Url}\fi

\bibitem[Gema et~al.(2025)Gema, Leang, Hong, Devoto, Mancino, Saxena, He, Zhao,
  Du, Madani, Barale, McHardy, Harris, Kaddour, van Krieken, and
  Minervini]{gema2025mmluredux}
Aryo~Pradipta Gema, Joshua Ong~Jun Leang, Giwon Hong, Alessio Devoto, Alberto
  Carlo~Maria Mancino, Rohit Saxena, Xuanli He, Yu~Zhao, Xiaotang Du, Mohammad
  Reza~Ghasemi Madani, Claire Barale, Robert McHardy, Joshua Harris, Jean
  Kaddour, Emile van Krieken, and Pasquale Minervini.
\newblock Are we done with mmlu?
\newblock In \emph{Proceedings of the 2025 Conference of the Nations of the
  Americas Chapter of the Association for Computational Linguistics: Human
  Language Technologies (Volume 1: Long Papers)}, pages 5069--5096, 2025.

\bibitem[Hendrycks et~al.(2021)Hendrycks, Burns, Basart, Zou, Mazeika, Song,
  and Steinhardt]{hendrycks2021mmlu}
Dan Hendrycks, Collin Burns, Steven Basart, Andy Zou, Mantas Mazeika, Dawn
  Song, and Jacob Steinhardt.
\newblock Measuring massive multitask language understanding.
\newblock In \emph{International Conference on Learning Representations}, 2021.

\bibitem[Hsu et~al.(2023)Hsu, Liu, Liao, Hsu, Chen, and Shiu]{hsu2023tceval}
Chan-Jan Hsu, Chang-Le Liu, Feng-Ting Liao, Po-Chun Hsu, Yi-Chang Chen, and
  Da-shan Shiu.
\newblock Advancing the evaluation of traditional chinese language models:
  Towards a comprehensive benchmark suite.
\newblock \emph{arXiv preprint arXiv:2309.08448}, 2023.

\bibitem[Huang et~al.(2023)Huang, Bai, Zhu, Zhang, Zhang, Su, Liu, Lv, Zhang,
  Lei, Fu, Sun, and He]{huang2023ceval}
Yuzhen Huang, Yuzhuo Bai, Zhihao Zhu, Junlei Zhang, Jinghan Zhang, Tangjun Su,
  Junteng Liu, Chuancheng Lv, Yikai Zhang, Jiayi Lei, Yao Fu, Maosong Sun, and
  Junxian He.
\newblock C-eval: A multi-level multi-discipline chinese evaluation suite for
  foundation models.
\newblock \emph{Advances in Neural Information Processing Systems}, 36, 2023.

\bibitem[Li et~al.(2024)Li, Zhang, Koto, Yang, Zhao, Gong, Duan, and
  Baldwin]{li2023cmmlu}
Haonan Li, Yixuan Zhang, Fajri Koto, Yifei Yang, Hai Zhao, Yeyun Gong, Nan
  Duan, and Timothy Baldwin.
\newblock Cmmlu: Measuring massive multitask language understanding in chinese.
\newblock In \emph{Findings of the Association for Computational Linguistics:
  ACL 2024}, pages 11260--11285, 2024.

\bibitem[Patil et~al.(2025)Patil, Mao, Yan, Ji, Suresh, Stoica, and
  Gonzalez]{patil2025bfcl}
Shishir~G. Patil, Huanzhi Mao, Fanjia Yan, Charlie Cheng-Jie Ji, Vishnu Suresh,
  Ion Stoica, and Joseph~E. Gonzalez.
\newblock The berkeley function calling leaderboard (bfcl): From tool use to
  agentic evaluation of large language models.
\newblock In \emph{Proceedings of the 42nd International Conference on Machine
  Learning}, volume 267 of \emph{Proceedings of Machine Learning Research},
  pages 48371--48392, 2025.

\bibitem[Sclar et~al.(2024)Sclar, Choi, Tsvetkov, and
  Suhr]{sclar2024quantifying}
Melanie Sclar, Yejin Choi, Yulia Tsvetkov, and Alane Suhr.
\newblock Quantifying language models' sensitivity to spurious features in
  prompt design or: How i learned to start worrying about prompt formatting.
\newblock In \emph{International Conference on Learning Representations}, 2024.

\bibitem[Tam et~al.(2024)Tam, Pai, Lee, Chen, Chu, Cheng, and
  Shuai]{tam2024tmmluplus}
Zhi-Rui Tam, Ya-Ting Pai, Yen-Wei Lee, Jun-Da Chen, Wei-Min Chu, Sega Cheng,
  and Hong-Han Shuai.
\newblock An improved traditional chinese evaluation suite for foundation
  models.
\newblock \emph{arXiv preprint arXiv:2403.01858}, 2024.

\bibitem[Zheng et~al.(2024)Zheng, Zhou, Meng, Zhou, and Huang]{zheng2024large}
Chujie Zheng, Hao Zhou, Fandong Meng, Jie Zhou, and Minlie Huang.
\newblock Large language models are not robust multiple choice selectors.
\newblock In \emph{International Conference on Learning Representations}, 2024.

\bibitem[Zheng et~al.(2023)Zheng, Chiang, Sheng, Zhuang, Wu, Zhuang, Lin, Li,
  Li, Xing, Zhang, Gonzalez, and Stoica]{zheng2023judging}
Lianmin Zheng, Wei-Lin Chiang, Ying Sheng, Siyuan Zhuang, Zhanghao Wu, Yonghao
  Zhuang, Zi~Lin, Zhuohan Li, Dacheng Li, Eric~P. Xing, Hao Zhang, Joseph~E.
  Gonzalez, and Ion Stoica.
\newblock Judging llm-as-a-judge with mt-bench and chatbot arena.
\newblock \emph{Advances in Neural Information Processing Systems}, 36, 2023.

\end{thebibliography}

\appendix
\section{Answer-Key Conflicts}
\label{app:conflicts}

Table~\ref{tab:conflicts} lists every one of the 27 duplicate groups (of
31{,}312) our detector flagged as answer-key disagreements across
printings, each manually attributed from its full printings. The
attribution matters more than the count: \textbf{14} depend on figures
absent from the extracted text (two different flow-cytometry plots, ECG
traces, or ventilator waveforms reusing the same stem --- the keys
``disagree'' because the pictures differ); \textbf{9} are detector
artifacts where every printing keys the same letter and the answer texts
differ only in whitespace or punctuation
(\zh{0℃為100Ω} vs.\ \zh{0℃為100 Ω}); \textbf{1} had its option set
legitimately rewritten between years so the key followed; \textbf{1} has a
deictic stem (\zh{該脊髓損傷患者}) pointing at a case vignette outside the
item; and only \textbf{2} are true conflicts --- self-contained, text-only
stems whose official keys genuinely disagree across years. The
figure-dependent and vignette groups are exactly what the filters of
Section~\ref{sec:filtering} remove; the two true conflicts put the
irreducible label noise of the source at $2/31{,}312 \approx 6 \times
10^{-5}$ per duplicate group, an order of magnitude below the conservative
detector-level figure quoted in the body.

% auto-generated by scripts/conflict_table.py from data/build-stats/conflicts.json
\begin{longtable}{@{}p{0.17\textwidth}p{0.33\textwidth}p{0.32\textwidth}p{0.09\textwidth}@{}}
\toprule
Subject & Question (truncated) & Official keys by year & Class \\
\midrule
\endhead
clinical-hematology & \zh{下圖是一位健康成年人的骨髓細胞以流式細胞儀（flow cytometry）分\ldots} & \zh{104: D (lymphocyte); 105: D (monocyte)} & figure \\
clinical-physiology-pathology & \zh{這張心電圖最恰當的診斷為何？\ldots} & \zh{105: A (第一度房室間傳導阻斷); 107: C (第三度房室間傳導阻斷)} & figure \\
critical-care-respiratory & \zh{先天性心臟病的兒童經過開心手術後，常有許多術後心臟功能異常，下列心血管壓力監\ldots} & \zh{103: A (左心室功能異常：右心房壓力正); 103: A (左心室功能異常：右心房壓力：)} & typo. \\
dental-technology & \zh{圖中標示星號（*）之肌肉是：\ldots} & \zh{101: A (顳肌（temporal mu); 101: D (咬肌（masseteric )} & figure \\
dental-technology & \zh{圖中，上顎齒列依殘留齒與缺牙區的位置關係為甘迺迪（Kennedy）分類法中之\ldots} & \zh{102: C (I 類-3 級); 102: D (I 類-4 級)} & figure \\
dentistry & \zh{圖中黑色箭頭所指的解剖位置為何？\ldots} & \zh{111: D (left mandible); 112: A (left maxilla)} & figure \\
freight-operations & \zh{依據國際海運危險品準則（IMDG CODE）規定，此標籤屬於何種類別之物質？\ldots} & \zh{101: B (放射性物質); 102: C (腐蝕性物質)} & figure \\
freight-operations & \zh{依據配艙法則，為滿足船舶具有適當穩度，對具有二層艙之船舶來說，其下大艙的貨重\ldots} & \zh{101: B (1/3); 103: D (2/3)} & \textbf{true} \\
legal-knowledge & \zh{關於婦女保障名額之規定，下列何者正確？\ldots} & \zh{106: B (我國各選舉區選出之直轄市議員)} & typo. \\
marine-electrical-control & \zh{於直流電路中有兩只同型電容器，其規格為500 V、0.1$\mu$F。今若予串聯，試\ldots} & \zh{102: C (1000 V、0.05$\mu$F); 103: C (1000 V，0.05$\mu$F)} & typo. \\
marine-electrical-control & \zh{使用磁脈衝感測器，其磁頭輸出的信號近似：\ldots} & \zh{102: C (正弦波); 103: A (方波)} & \textbf{true} \\
marine-engine-maintenance & \zh{PT100之溫度量測元件，在檢驗好壞時，可使用歐姆計量測其電阻，其基準電阻為\ldots} & \zh{101: B (0℃為100Ω); 104: B (0℃為100 Ω)} & typo. \\
maritime-communications & \zh{下圖為救援單位指引遇險船舶人員登岸之信號，其意義為：\ldots} & \zh{101: B (此處登岸極危險); 101: A (這是最佳登岸地點) \ldots} & figure \\
maritime-communications & \zh{下圖為搜救船舶回答搜救飛機之信號，其意義為：\ldots} & \zh{101: C (收到信號，無法配合); 101: B (收到信號，可配合)} & figure \\
mechanical-ventilation & \zh{下列圖形顯示病人使用的通氣模式為何？\ldots} & \zh{102: B (壓力目標控制型通氣（pres); 103: A (容積目標控制型通氣（volu) \ldots} & figure \\
mechanical-ventilation & \zh{病人使用呼吸器顯示下列圖形時，何種處理方式最正確？\ldots} & \zh{102: D (尋找漏氣的原因); 102: D (氣道分泌物的抽吸) \ldots} & figure \\
mechanical-ventilation & \zh{病人使用容積控制型通氣模式，恆定吸氣流速，病人生理參數如下，下列敘述何者正確\ldots} & \zh{105: B (僅\cnum{2}\cnum{3}\cnum{6}); 106: C (僅\cnum{2}\cnum{4}\cnum{5})} & revised \\
mechanical-ventilation & \zh{下圖為何種通氣模式？\ldots} & \zh{106: B (容積控制型通氣（volume); 106: A (壓力控制型通氣（pressu)} & figure \\
molecular-diagnostics & \zh{下圖為何種儀器？\ldots} & \zh{102: A (尿液比重計（urinomet); 104: B (折射儀（refractome)} & figure \\
neuro-physical-therapy & \zh{該脊髓損傷患者符合那種臨床症候群？\ldots} & \zh{112: C (布朗－斯夸症候群); 114: B (中髓症候群)} & vignette \\
occupational-therapy & \zh{人類職能模式（ Model of Human Occupation） 認為人\ldots} & \zh{101: B (職能參與（ occupati); 105: B (職能參與（occupatio)} & typo. \\
occupational-therapy-techniques & \zh{有關射飛鏢活動之分析，下列敘述何者錯誤？\ldots} & \zh{101: C (Biceps 是射出飛鏢的重); 105: C (Biceps是射出飛鏢的重要)} & typo. \\
pharmaceutical-analysis & \zh{請問下圖為何種儀器的裝置示意圖？\ldots} & \zh{104: B (螢光光譜儀); 105: C (紫外光－可見光光譜儀)} & figure \\
physical-dysfunction-ot & \zh{下列何項感覺測試，不適合用來評估脊髓損傷的受傷層級？\ldots} & \zh{101: A (觸辨覺（ stereogno); 106: A (觸辨覺（stereognos)} & typo. \\
radiographic-equipment & \zh{加馬攝影機中，下列有關其輸出信號的脈衝振幅的敘述，何者正確？\ldots} & \zh{101: C (與撞擊NaI(Tl)晶體的加); 104: C (與撞擊NaI（Tl）晶體的加)} & typo. \\
respiratory-diseases & \zh{下列ECG是那一種心律不整？\ldots} & \zh{101: D (Paroxysmal sup); 103: C (Atrial fibrill)} & figure \\
veterinary-pharmacology & \zh{何者為 enalapril 擴張血管的主要作用機制？\ldots} & \zh{101: D (抑制 angiotensin); 104: D (抑制angiotensin-)} & typo. \\
\bottomrule
\caption{All 27 flagged answer-key conflict groups with manual attribution. \emph{figure}: stem names a figure the text does not carry; \emph{typo.}: printings differ only in whitespace or punctuation (detector artifact, not a conflict); \emph{revised}: option set legitimately rewritten between years; \emph{vignette}: deictic stem pointing at case context outside the item; \emph{true}: self-contained stem whose official keys genuinely disagree.}
\label{tab:conflicts}
\end{longtable}

\section{TMMLU+ Overlap Lists}
\label{app:tmmlu}

We ran the same duplicate audit on TMMLU+ that Section~\ref{sec:keys} runs
on our own source, grouping byte-identical items (NFKC-normalised stem plus
the \emph{sorted} option multiset, so that a re-printing that shuffles the
letters is the same item and a letter change alone is not a conflict).
Across its 23{,}072 rows: \textbf{821} duplicate groups covering 1{,}720
items; exactly \textbf{one} true answer-key conflict --- a polymer-counting
chemistry item keyed once as ``5'' and once as ``7''; \textbf{431} groups
recur across different splits (899 items); and \textbf{136} of the 20{,}160
test items appear verbatim in a non-test split (train, val, or the few-shot
dev pool), including 22 in train itself. The label accuracy is excellent --- one conflict in 821 groups
($0.12\%$), the same order of magnitude as our source's 27 in 31{,}312
($0.09\%$) --- but the cross-split recurrence means few-shot evaluation on
TMMLU+ can show a model its own test items as exemplars. The audit script
and machine-readable lists are released alongside the datasets
(\texttt{scripts/audit\_tmmluplus.py},
\texttt{data/build-stats/tmmluplus\_overlap.json}).

\section{Prompts}
\label{app:prompts}

\paragraph{Evaluation system prompt.} Both versions differ only in the
sentence that names the answer format; the defective version's example
letter \zh{甲} is what Section~\ref{sec:placeholder} traces the Mistral
collapse to.

\begin{quote}\small
\textbf{Defective (with placeholder example):}\\
\zh{你是一位專業的測驗作答助理。請仔細閱讀題目，以繁體中文（臺灣用語）思考，
並在最後將唯一正確選項的英文字母放入 \texttt{\textbackslash boxed\{\}} 中，
例如 \texttt{\textbackslash boxed\{\zh{甲}\}} 的位置應填入選項字母。
除了 \texttt{\textbackslash boxed\{\}} 之外不要使用其他格式標示答案。}

\textbf{Final:}\\
\zh{你是一位專業的測驗作答助理。請仔細閱讀題目，以繁體中文（臺灣用語）思考，
並在最後輸出 \texttt{\textbackslash boxed\{\}}，大括號內只填入唯一正確選項的
英文字母，也就是 A、B、C、D 其中一個。除了 \texttt{\textbackslash boxed\{\}}
之外不要使用其他格式標示答案。}
\end{quote}

\paragraph{Formosa generation prompt (verbatim).}

\begin{quote}\small
\zh{你是臺灣的測驗命題委員，正在編寫一份「認識臺灣」的知識評測集。
受測者只會看到題目，看不到任何參考資料，必須靠自己對臺灣的認識作答。}

\zh{短文只是你的取材來源，不是題目的一部分。請從短文中挑出\textbf{值得知道的
臺灣知識}來出題。}

\zh{必須遵守：}

\begin{sloppypar}
\zh{1. 題幹要自己說得清楚。人名、地名、機構名、作品名一律寫出來。
絕對不可以用「他」「她」「它」「該」「此」「這位」來指涉短文裡的人事物。
反例（禁止）：「她取得美國國籍的年份是？」
正例：「作家陳若曦取得美國國籍是在哪一年？」}
\end{sloppypar}

\zh{2. 絕對不可以出現「短文」「本文」「文中」「上文」「文章」
「根據以上」等字眼，也不可以用「製作說明中」「文件裡」
這種暗指來源的說法。}

\zh{3. 只出在其他地方也查得到的臺灣知識 —— 歷史事件、人物事蹟、地理、制度法規、
文化習俗、著名作品、產業發展、自然生態。}

\zh{4. 不要出只有這份文件才知道的細節：報名費與價格、某一屆活動的數字、
網站或專案的內部統計、功能上線或下架的日期、單一公司的內部流程、
某個專案有幾個圖層。自問：「一個很懂臺灣但沒讀過這份文件的人，
有沒有可能知道答案？」若不可能，換一題。}

\zh{5. 三個誘答選項要合理但明確錯誤，長度與正確答案相近。}

\zh{6. 一律使用臺灣用語與繁體中文（「網路」而非「網絡」、「軟體」而非「軟件」）。}

\zh{7. 題幹不可直接出現答案文字。}
\end{quote}

\paragraph{Grounding-judge prompt (verbatim).}

\begin{quote}\small
\zh{你是臺灣測驗題目的審題委員。你會看到一段來源文字和一道依據它出的選擇題。}

\zh{請判斷兩件事：}

\zh{--- supported：標示為正確的那個選項，是否真的能從來源文字中找到依據？
改寫、摘要、合併來源中的多個事實都算有依據。若該選項的內容在來源裡找不到，
或來源其實支持別的選項，就是 false。}

\zh{--- single\_answer：是否只有一個選項正確？若有第二個選項也說得通，
就是 false。}

\zh{只輸出 JSON：\{"supported": true, "single\_answer": true,
"reason": "一句話"\}}
\end{quote}

\section{Reproducibility}
\label{app:repro}

All evaluations use Twinkle Eval's \texttt{box} strategy with temperature 0,
a 16{,}384-token generation budget for the full-corpus exam run, and
published parse rates per cell. The Formosa curation manifest freezes the
pool membership, band, and date; per-item difficulty sidecars allow any
alternative band to be recomputed. Table~\ref{tab:repro} lists every
evaluation run behind a number in this paper.

\begin{table}[h]
\centering
\small
\begin{tabular}{llll}
\toprule
Run & Models & Serving & Dates (2026) \\
\midrule
\exambench{} full corpus & Qwen3.8-27B & self-hosted vLLM & Aug 22--24 \\
\exambench{} budget study & Qwen3.8-27B & self-hosted vLLM & Aug 22 \\
Formosa pool (7 models) & NCHC catalog (see caption) & NCHC API & Aug 23 \\
Formosa selection frozen & --- & manifest v2.1 & Aug 24 \\
Formosa held-out & Qwen3.8-27B & self-hosted vLLM & Aug 24 \\
Formosa generation & gpt-oss-120b & NCHC API & Aug 23 \\
Judge calibration & 10 candidates & NCHC API & Aug 23 \\
\bottomrule
\end{tabular}
\caption{Runs, serving paths, and dates. The frozen pool is
Mistral-Large-3-675B-Instruct-2512, Llama-3.1-405B-Instruct-FP8,
gemma-4-31B-it, NVIDIA-Nemotron-3-Super-120B-A12B, Gemma-3-27B,
gpt-oss-120b, and Llama-3.3-70B-Instruct, all served by Taiwan's National
Center for High-performance Computing (NCHC). All runs use Twinkle Eval's
\texttt{box} strategy, temperature $0$, \texttt{top\_p} $0.9$,
\texttt{shuffle\_options} on; generation caps are 16{,}384 tokens for the
full-corpus exam run and 4{,}096 for all Formosa runs. Datasets, per-item
difficulty sidecars, the curation manifest, per-cell parse rates, and the
build manifests are published with the benchmark repositories.}
\label{tab:repro}
\end{table}

\end{document}